\documentclass[11pt,a4paper]{article}
\usepackage[hyperref]{acl2021}
\usepackage{times}
\usepackage{latexsym}

\usepackage{graphicx}
\usepackage{amsmath}
\usepackage{booktabs}
\usepackage{multirow}
\usepackage{comment}
\usepackage{bm}
\usepackage{microtype}
\usepackage{amssymb}
\usepackage{pifont}
\newcommand{\cmark}{\ding{51}}%
\newcommand{\xmark}{\ding{55}}%

\aclfinalcopy 

\title{
Seeing past words: Testing the cross-modal capabilities of\\pretrained V\&L models on counting tasks} 

\author{Letitia Parcalabescu$^1$\ \ ~~Albert Gatt$^2$\ \ ~~Anette Frank$^1$\ \ ~~Iacer Calixto$^{3,4}$\\\\
$^1$Heidelberg University, Department of Computational Linguistics\\ 
$^2$University of Malta, Institute of Linguistics and Language Technology \\
$^3$New York University \ \ 
$^4$ILLC, University of Amsterdam\\
\texttt{\{parcalabescu,frank\}@cl.uni-heidelberg.de}\\ \texttt{albert.gatt@um.edu.mt, iacer.calixto@nyu.edu}
}

\date{}

\begin{document}
\maketitle
\begin{abstract}
We investigate the reasoning ability of pretrained vision and language (V\&L) models in two tasks that require multimodal integration: (1) discriminating a correct image-sentence pair from an incorrect one, and (2) counting entities in an image.
We evaluate three pretrained V\&L models on these tasks: ViLBERT, \mbox{ViLBERT 12-in-1} and LXMERT, in  zero-shot and finetuned settings.
Our results show that models solve task (1) very well, as expected, since all models are pretrained on task (1).
However, none of the pretrained V\&L models is able to adequately solve task (2), our counting probe, and they cannot generalise to out-of-distribution quantities. 
We propose a number of explanations for these findings: LXMERT (and to some extent \mbox{ViLBERT 12-in-1}) show some evidence of catastrophic forgetting on task (1).
Concerning our results on the counting probe, we find evidence that all models are impacted by dataset bias, and also fail to individuate entities in the visual input.
While a selling point of pretrained V\&L models is their ability to solve complex tasks, our findings suggest that understanding their reasoning and grounding capabilities requires more targeted investigations on specific phenomena. 
\end{abstract}

\section{Introduction}
Recently, many vision and language (V\&L) models that combine images and text have been proposed \cite{lu2019vilbert,tan-bansal-2019-lxmert,li2019visualbert,chen2020uniter,Li-etal-2020unicodervl,Su2020VL-BERT}.
These models follow the \textit{pretrain-and-finetune} paradigm,
i.e. they are pretrained using self-supervision on large amounts of image-caption pairs\footnote{Sometimes models are also pretrained on other image-text datasets, e.g., visual question answering data.} and are then finetuned on the task(s) of interest.
Such V\&L models have obtained state-of-the-art performance across a number of different V\&L tasks, e.g. visual question answering (VQA); visual commonsense reasoning; grounding referring expressions; and image retrieval, among others.

Pretrained V\&L models use a combination of masked multimodal modelling -- i.e., masking out words and object bounding boxes from the input and predicting them -- and image-sentence alignment, i.e., predicting whether an image-sentence pair is correctly aligned or not. Such models hold the promise of partially addressing the `meaning gap' in unimodal pretrained language models such as BERT \cite{devlin2019bert} by directly connecting language to visual representations \cite{Bender2020,bisk-etal-2020-experience}.

In this paper, we use foiling to investigate how well pretrained V\&L models integrate and reason upon textual and visual representations.
The foiling strategy, introduced by \citet{shekhar-etal-2017-foil} in the context of vision and language tasks, relies on replacing an element in a text with another element, such that the replacement results in a mismatch with the image.
We propose two tasks which require effective multimodal integration:
(1) discriminating a correctly aligned image-sentence pair from an incorrectly aligned one, and (2) \textit{counting} entities in the image. 

V\&L models are commonly pretrained on task (1), and should not have many difficulties detecting incorrect image-sentence pairs.
Counting, our task (2), nicely puts together visual and textual reasoning. It requires the detection of \textit{object instances} in the visual input,
mapping these instances to \textit{categories}, as well as properly aligning such instances to references in the textual input.
Model architectures have been proposed \textit{specifically} for counting, which is known to be a hard V\&L problem \cite{zhang2018learning,Acharya_Kafle_Kanan_2019TallyQA,trott2018interpretable,Chattopadhyay-etal-2017}. Unlike these specialised approaches, we focus on general-purpose V\&L models. 
Related V\&L work has also investigated generalised quantifiers (such as \textit{most}) in a V\&L context, but this work has generally exploited synthetic datasets \cite{Sorodoc2018,pezzelle-fernandez-2019-red,Testoni2019}.
Here, we task the model to judge whether an unambiguous question or statement about \textit{the number of entities visible in a natural image} is correct.

We use three publicly available, representative V\&L models in our investigation: 
LXMERT\footnote{\url{github.com/huggingface/transformers}} \cite{tan-bansal-2019-lxmert}, ViLBERT and ViLBERT 12-in-1\footnote{\url{github.com/facebookresearch/vilbert-multi-task}} \citep[][]{lu2019vilbert,lu2020vilbert12in1}.
ViLBERT and ViLBERT 12-in-1 use the same BERT-based model architecture, which incorporates two separate visual and linguistic streams that interact through multiple co-attention transformer layers. 
ViLBERT is trained using self-supervised learning on image-caption pairs, while ViLBERT 12-in-1 is further finetuned on 12 different tasks using multi-task learning. LXMERT is also a dual-stream architecture and combines textual and visual transformer-based encoders with cross-modal layers. 
However, LXMERT is pretrained not only on image-caption pairs but also directly on the visual question answering (VQA) task using multiple VQA datasets.

While all models are trained on image-sentence alignment, only ViLBERT is not directly trained on VQA; hence the model can be probed ``zero-shot'' on our counting task.
LXMERT, by contrast, is pretrained on VQA (including \textit{how many} questions, the focus in our counting probe). Hence, LXMERT was exposed to examples where answering a question correctly
requires the model to detect and categorise instances in an image, and then aligning these to the text.
Finally, the tasks \mbox{ViLBERT 12-in-1} are finetuned on also include VQA, including instances with numerical answers requiring counting abilities.
We therefore believe it serves as a solid baseline and should be well equipped to detect foiled probing instances. To our surprise, we find that none of these models perform particularly well in our counting experiments.


Our main contributions are:
i) We show that all three models perform image-sentence alignment well, as expected given their pretraining;
ii) We build a \textit{counting probe}, which requires a model to  adequately perform cross-modal grounding;
iii) We find that ViLBERT, \mbox{ViLBERT 12-in-1} and LXMERT perform similarly to the random baseline when directly applying the image-sentence alignment head to perform counting without finetuning;
iv) We find that all models seem to exploit dataset bias and fail to generalise to out-of-distribution quantities. Even when finetuned, they only partially solve our counting probe.\footnote{We will release all data necessary to reproduce our experiments, including our counting dataset, upon publication.}

\section{V\&L Models}\label{sec:vision_and_language_models}
The pretrained V\&L models we use are {\bf ViLBERT} \citep{lu2019vilbert}, {\bf ViLBERT 12-in-1} \citep{lu2020vilbert12in1}, and {\bf LXMERT} \citep{tan-bansal-2019-lxmert}.
ViLBERT is pretrained on  Google Conceptual Captions \citep{sharma2018conceptual} on (i) multimodal masked prediction, i.e., masking objects and words and predicting them; and (ii) image-sentence alignment, i.e., determining whether a text corresponds to an image or not.
LXMERT uses losses (i) and (ii) and is additionally pretrained on multiple VQA datasets, as well as object labelling.
Finally, ViLBERT 12-in-1 starts from a pretrained ViLBERT model checkpoint and is additionally finetuned on 12 different tasks, once again including VQA.

\subsection{Evaluation}\label{sec:evaluation_metrics}
In both image-sentence alignment and counting probes, models are exposed to either correct or foiled image-text pairs. We evaluate pretrained V\&L models on our probes using {\bf accuracy} ($acc$), which is the overall accuracy on all classes; {\bf precision} ($p_c$), which measures how well the models identify the \textit{correct} examples; and {\bf foil precision} ($p_f$), which measures how well a model identifies \textit{foiled} instances:
\begin{equation*} \label{eq:metrics}
    \begin{split}
       acc &= \frac{P + N}{P + N + \tilde{P} + \tilde{N}} , \\
       p_c &= P / (P + \tilde{P}) , \\
       p_f &= N / (N + \tilde{N}) , 
  \end{split}
\end{equation*}
where $P$ and $N$ are the number of true positives and true negatives, and $\tilde{P}$ and $\tilde{N}$ are the number of false positives and false negatives, respectively.

We also evaluate our models using a {\bf pairwise ranking accuracy} $acc_r$ computed using the image-sentence alignment score $\phi$ that the model assigns to correct and foiled image-text pairs.
Given an image ($i$) paired with a correct ($c$) versus a foil ($f$) text, if the score of the positive/correct pair is greater than that of the foiled pair, the prediction is considered successful.
\begin{equation*} \label{eq:metric_pairwise}
    \begin{split}
       acc_r &= \frac{\sum_{(i,c) \in C} \sum_{f \in F} s(i,c,f)}{|C| + |F|}, \\
       s(i,c,f) &=
       \begin{cases}
           1, & \text{if } \phi(i, f) \le \phi(i, c),\\
           0, & \text{otherwise,}
       \end{cases}
    \end{split}
\end{equation*}
\noindent
where $C$ is the set of correct image-caption pairs ($i,c$), and $F$ is the set of foils for the pair ($i,c$).

\section{Image-Sentence Alignment Probe} \label{sec:image_sentence_alignment}
In this set of experiments, we probe whether pretrained V\&L models can distinguish correct image-sentence pairs from foiled ones. While all models under consideration have been pretrained on this task, results are not usually reported for pretraining. Our aim is to explicitly establish their capabilities on a fundamental V\&L task that we would expect them to perform well at, before probing them on the more challenging counting task.

\subsection{Data}
To probe our models on the image-sentence alignment task, we construct evaluation sets using 5000 images each from the MSCOCO \cite{Lin-etal:2014:mscoco} and Google Conceptual Captions \cite[GCC;][]{sharma2018conceptual} validation splits.
MSCOCO images are collected from Flickr and its captions are crowdsourced.
GCC's images are obtained from the web with captions harvested from online alt-text enabled sources, and therefore contain more noise and variability.
ViLBERT is pretrained on GCC image-caption pairs; LXMERT is pretrained on five datasets including MSCOCO, but not GCC.
For both datasets we select one correct caption for each image, and create foils by pairing the image to one random caption from the remaining 4999 images. All models are tested on the same data.

\subsection{Experiments}\label{sec:ism_results}

In these experiments, we probe pretrained V\&L models without any additional fine-tuning.
Table~\ref{tab:coco_gcc_zeroshot} reports the results of applying the models' pretrained image-sentence alignment prediction head to image-caption pairs from MSCOCO and GCC.
We also highlight which models can be considered ``zero-shot'' in this setting:
GCC is used to pretrain the two ViLBERT models, while MSCOCO is used when pretraining LXMERT.



\begin{table}[t!]
    \small
    \centering
    \begin{tabular}{llc rrr}
        & \bf Model & ZS? & $acc$ & $p_c$ & $p_f$ \\
        \toprule
        & Random & & 50.0 & 50.0 & 50.0 \\
        \midrule
        \parbox[t]{2mm}{\multirow{3}{*}{\rotatebox[origin=c]{90}{\bf COCO}}}
        & ViLBERT & \cmark{} & 97.4 & 98.0 & 96.8 \\
        & ViLBERT 12-in-1 & \cmark{} & 96.4 & 93.4 & 99.4 \\
        & LXMERT & \xmark{} & 85.5 & 71.5 & 99.6 \\

        \midrule
        \parbox[t]{2mm}{\multirow{3}{*}{\rotatebox[origin=c]{90}{\bf GCC}}}
        & ViLBERT & \xmark{} & 96.8 & 96.7 & 96.9 \\
        & ViLBERT 12-in-1 & \xmark{} & 84.9 & 73.1 & 96.7 \\
        & LXMERT & \cmark{} & 67.9 & 31.9 & 97.9 \\
        \bottomrule
    \end{tabular}
    \caption{Image-sentence alignment results on our COCO and Google CC validation sets. 'ZS?' indicates whether the model is applied zero-shot, i.e., the model was never trained on examples from MSCOCO/GCC.
    We report the overall accuracy $acc$, precision on correct examples $p_c$, and precision on foiled examples $p_f$.
    }
    \label{tab:coco_gcc_zeroshot}
\end{table}

ViLBERT performs very well on both datasets and achieves $96$--$97$ $acc$ overall.
It predicts both correct and foiled examples well, as shown by $96$--$98$ $p_c$ and $\sim 96$ $p_f$.
When using ViLBERT 12-in-1, results on GCC are considerably worse compared to ViLBERT.
This is surprising, since ViLBERT 12-in-1 was trained using more tasks and considerably more data than ViLBERT.
Finally, LXMERT performs worst overall among all three models.

These results suggest that LXMERT (and to a lesser extent, ViLBERT 12-in-1) may be exhibiting catastrophic forgetting, a well-studied problem in neural networks \cite{Robins1995} which has received attention in NLP \cite{Kirkpatrick2017,yogatama2019learning} as well as in V\&L tasks in particular \cite{Greco2019}:
LXMERT is finetuned on visual question answering in the last 10 epochs of pretraining, and ViLBERT 12-in-1 is finetuned on 12 different tasks.
This finetuning may be responsible for the worse results observed, resulting in a downgrading of performance on the task the models were originally pretrained on.

In summary, all models solve the image-sentence alignment probe well (as expected) but the models show notable differences in performance; we conjecture catastrophic forgetting may be impacting the finetuning procedure of each model differently.

\section{Counting Probe}\label{sec:counting}

\begin{figure}
    \centering
    \includegraphics[width=.5\textwidth]{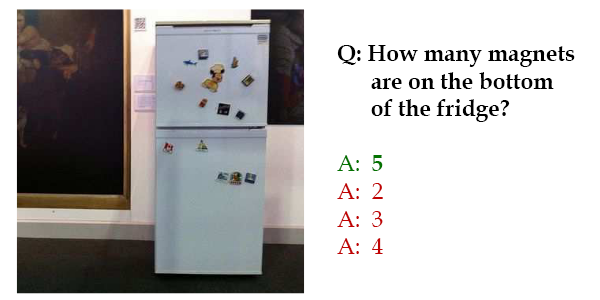}
    \caption{\textit{How many} question from Visual7W dataset.}
    \label{fig:visual7w_example}
\end{figure}

In our second task we probe  pretrained V\&L models on their ability to \textit{count}, i.e., to correctly predict the number of entities visible in an image,  
given the image itself and either a corresponding question coupled with a numerical answer, or a declarative statement about the number of entities of a specific kind derived from the question-answer pair (see Figure~\ref{fig:visual7w_example}).

\subsection{Data}

We collect our counting probe data from Visual7W \citep{zhu2016cvpr}, a VQA dataset with diverse question types including \textit{how many} questions, where a correct answer requires the model to count the number of entities of a certain type in an image.

\subsubsection{Data Formats}
The data is originally in question-answering format, where the answer is a number. We experiment with two alternative formats.

\paragraph{Q+A format}
We concatenate the original question with the separator token \texttt{[SEP]} and each answer (correct and foil), e.g. the example in Fig.\ \ref{fig:visual7w_example} becomes ``How many magnets are on the bottom of the fridge? \texttt{[SEP]} {\bf 2}/{\bf 3}/{\bf 4}/{\bf 5}''.

\paragraph{Declarative format}
ViLBERT is never pretrained on questions and answers with a separator token \texttt{[SEP]}.
We therefore create a version of the counting data where we transform the question and answer into a declarative statement using simple templates, described in detail in Appendix \ref{sec:appendix:counting_template}.
For instance, we create the following statements for the example shown in Fig.~\ref{fig:visual7w_example}: ``There are {\bf 2}/{\bf 3}/{\bf 4}/{\bf 5} magnets on the bottom of the fridge.''
Examples which could not be converted were removed.

\subsubsection{Data Splits}
To avoid leaks, instances extracted from a given Visual7W split are put into the same split in our counting dataset.\footnote{ I.e., V7W train $\rightarrow$ counting train, V7W valid $\rightarrow$ counting valid, V7W test $\rightarrow$ counting test.}

\begin{figure}[t]
    \centering
    \includegraphics[width=\columnwidth]{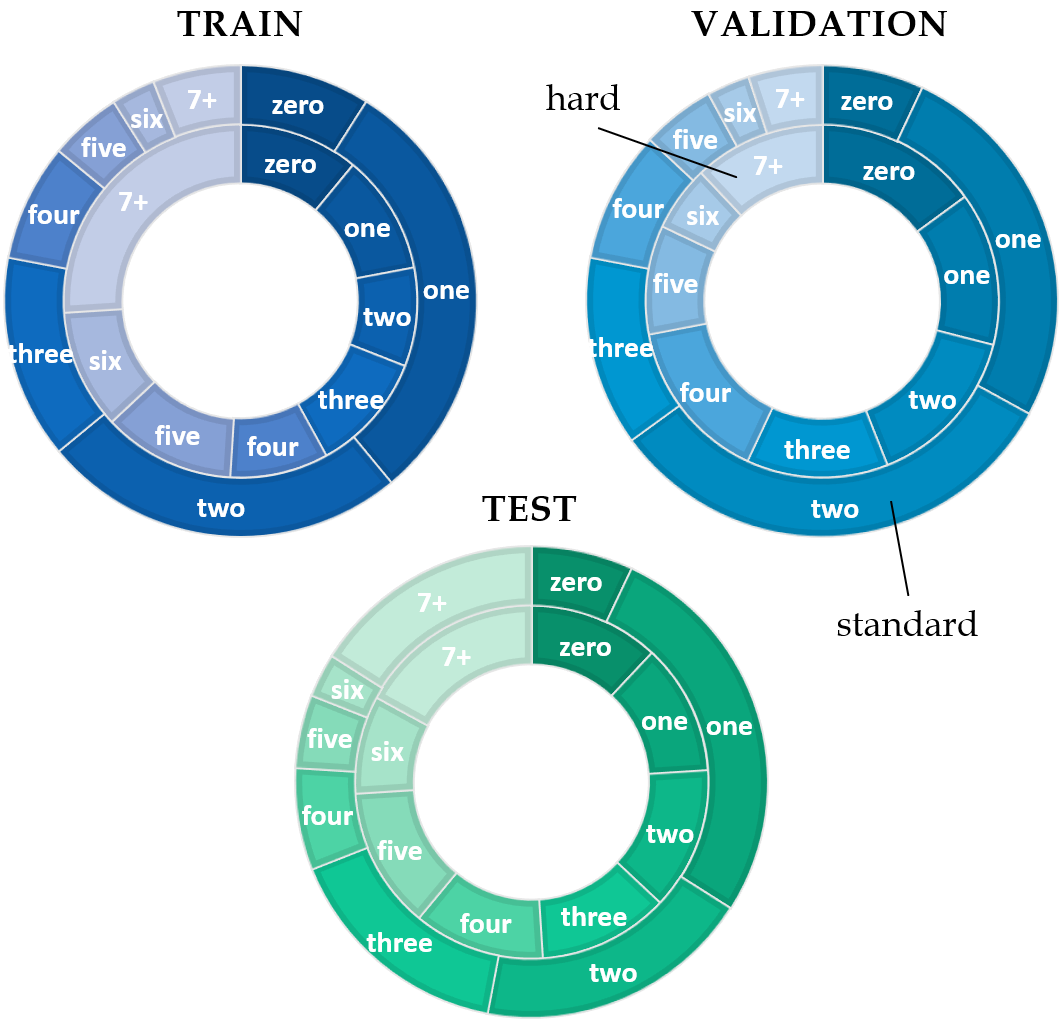}
    \caption{Percentage of numerals in the counting data. Outer circle: standard split, inner circle: hard split.}
    \label{fig:counting_data_statistics}
\end{figure}

We create three splits for our counting dataset: {\bf standard}, {\bf hard}, and {\bf interpolated}.
In the \textit{standard} split we include all examples of \textit{how many} questions in the train, dev and test splits in Visual7W, excluding examples that cannot be transformed into a declarative statement with our templates.
The distribution of numerals in the standard split is highly skewed and answers such as ``1'' or ``2'' are by far the most common (see the outer circles in Fig. \ref{fig:counting_data_statistics}).
We mitigate this by introducing a \textit{hard} split (see the inner circles in Fig. \ref{fig:counting_data_statistics}), in which high-frequency classes are capped at $k=200$ examples for train, dev and test sets, and any training examples where the answer is a number greater than 20 are removed.
Finally, in the \textit{interpolated} setting we split the original data so that only examples whose answers are even are in the training set, with validation and test sets only containing examples with odd answers.

Data statistics are reported in full in Appendix \ref{sec:appendix:counting_data}. We note that both the capping in the \textit{hard} split, and the interpolation in the \textit{interpolated} split, result in fewer instances.
The \textit{hard} split is more balanced with regards to the number of classes, whereas quantities in the \emph{standard} split follow a more natural distribution, where numerals like ``one'', ``two'' or ``three'' are more common than large quantities or mentions of empty sets (Figure~\ref{fig:counting_data_statistics}).
The less skewed distribution in the \emph{hard} split would be expected to be harder, since we artificially lower the relative frequency of frequent answers.

\subsection{Experiments}\label{sec:experiments}

We conduct a number of experiments where counting capabilities are probed in different ways, via \emph{image-sentence alignment} (Section \ref{sec:counting_as_image_sentence_alignment}), \emph{masked language modelling} (MLM; Section \ref{sec:MLM}), and \emph{visual question answering} (Section \ref{sec:LXMERT_VQA}).

It is important to note that there is a difference between the three models in terms of their prior exposure to the VQA task in general, and to questions involving counting in particular. Specifically, while ViLBERT was exclusively pretrained on GCC, LXMERT's pretraining involved VQA in the final ten epochs, and this included the Visual7W training set. In the case of ViLBERT 12-in-1, VQA was also one of the tasks on which it was finetuned, and again this included Visual7W.
In the experiments reported below, we distinguish between a {\em no finetuning} and a {\em finetuned} scenario. In the former, we present results on models which were {\em not directly finetuned on our counting training set}, irrespective of whether they were exposed to Visual7W during pretraining (as in the case of LXMERT) or training (as in the case of ViLBERT 12-in-1).
In the \textit{finetuned} scenario, we finetune each model using three different random seeds and report mean and standard deviation for all metrics.

\begin{table}[t!]
    \small
    \centering
    \begin{tabular}{ll rrrr}
        \bf Split & \bf Format & $acc$ & $p_c$ & $p_f$ & $acc_r$ \\
        \midrule
        \multicolumn{2}{l}{Random baseline} & 50.0 & 50.0 & 50.0 & 50.0 \\
        \midrule
        
        \multicolumn{6}{c}{ViLBERT}\\
        \midrule
        \multirow{2}{*}{std.} & Q+A & 37.8 & 74.3 & 25.5 & 49.0 \\
        & decl. & 37.6 & 77.9 & 24.1 & 57.0 \\
        \cmidrule{2-6}
        \multirow{2}{*}{hard} & Q+A & 38.6 & 73.1 & 27.1 & 51.9 \\
        & decl. & 38.0 & 75.3 & 25.5 & 55.9 \\
        \midrule
        
        \multicolumn{6}{c}{LXMERT}\\
        \midrule
        \multirow{2}{*}{std.} & Q+A & 50.5 & 45.7 & 55.2 & 57.2\\
        & decl. & 54.7 & 51.0 & 58.5 & 72.8 \\
        \cmidrule{2-6}
        \multirow{2}{*}{hard} & Q+A & 50.4 & 47.4 & 53.4 & 59.3 \\
        & decl. & 52.3 & 50.5 & 54.2 & 64.4 \\
        \midrule
        \multicolumn{6}{c}{ViLBERT 12-in-1}\\
        \midrule
        \multirow{2}{*}{std.} & Q+A & 43.3 & 80.2 & 30.9 & 77.3 \\
        & decl. & 62.4 & 73.7 & 58.7 & 75.4 \\
        \cmidrule{2-6}
        \multirow{2}{*}{hard} & Q+A & 46.9 & 67.1 & 40.1 & 70.3 \\
        & decl. & 61.3 & 70.0 & 58.3 & 72.6 \\
        
        \bottomrule
    \end{tabular}
    \caption{Counting: test results for models \textit{without} fine-tuning on our counting dataset, including ViLBERT ``zero-shot''. We report overall accuracy $acc$, precision on correct examples $p_c$, precision on foiled examples $p_f$, and pairwise accuracy $acc_r$. Splits: standard (std.) and hard. Formats: Q+A and declarative (decl).}
    \label{tab:counting_as_image_sentence_aligment_nofinetuning}
\end{table}

\subsubsection{Counting as Image-Sentence Alignment}\label{sec:counting_as_image_sentence_alignment}

\begin{table*}[t!]
    \small
    \centering
    \resizebox{\linewidth}{!}{
    \begin{tabular}{lr rrrr c rrrr}
        \toprule
        \bf Split & \bf Format & $acc$ & $p_c$ & $p_f$ & $acc_r$ && $acc$ & $p_c$ & $p_f$ & $acc_r$ \\
        \midrule
        \multicolumn{2}{l}{Random baseline} & 50.0 & 50.0 & 50.0 & 50.0  && 50.0 & 50.0 & 50.0 & 50.0 \\
        \midrule
        
        &&\multicolumn{4}{c}{ViLBERT} && \multicolumn{4}{c}{ViLBERT 12-in-1}\\
        \cmidrule{3-11}
        
        \multirow{2}{*}{std.} & Q+A & 74.4 (0.2) & 49.3 (0.3) & 88.9 (0.3) & 78.2 (0.6) && 81.1 (0.3) & 60.3 (0.7) & 90.2 (0.2) & 83.5 (0.1) \\
        & decl. & 71.7 (3.6) & 46.3 (4.1) & 88.6 (1.1) & 76.7 (2.7) && 81.1 (0.1) & 60.6 (0.1) & 89.7 (0.1) & 83.3 (0.1) \\
        \cmidrule{3-11}
        \multirow{2}{*}{hard} & Q+A & 56.7 (22.2) & 16.9 (11.9) & 75.4 (0.6) & 56.2 (0.7) && 64.3 (4.2) & 38.7 (3.3) & 86.0 (2.7) & 69.8 (2.6) \\
        & decl. & 54.0 (21.4) & 38.6 (14.4) & 52.3 (37.0) & 57.5 (0.9) && 71.9 (1.9) & 46.4 (1.9) & 89.4 (0.8) & 77.5 (0.6) \\
        \cmidrule{3-11}
        \multirow{2}{*}{interp.} & Q+A & 48.0 (0.2) & 0.2 (0.1) & 65.6 (0.1) & 12.8 (0.3)  && 52.5 (0.5) & 0.1 (0.1) & 67.6 (0.2) & 11.4 (1.4) \\
        & decl. & 49.1 (0.6) & 0.3 (0.2) & 66.2 (0.3) & 17.9 (0.8) && 52.7 (0.3) & 0.0 (0.0) & 67.7 (0.1) & 13.5 (2.8) \\
        
        \bottomrule
    \end{tabular}
    }
    \caption{Counting: test results for models fine-tuned on our counting training data. We report mean (std) over three runs: overall accuracy $acc$, precision on correct examples $p_c$, precision on foiled examples $p_f$, and pairwise accuracy $acc_r$. Splits: standard (std.), hard, and interpolated (interp.). Formats: Q+A and declarative (decl).}
    \label{tab:counting_as_image_sentence_aligment_finetuned}
\end{table*}

In this setup, we frame the counting task as an image-sentence alignment problem.
We use the pretrained V\&L models' image-sentence alignment head either to predict whether the sentence (in Q+A or declarative format) matches the image or not (i.e., in a per-example comparison evaluated with $acc$, $p_c$, $p_f$), or to score correct and foiled pairs (i.e., in a pairwise comparison evaluated with $acc_r$).
See Section \ref{sec:evaluation_metrics} for details on how we compute these metrics.
In Tables  \ref{tab:counting_as_image_sentence_aligment_nofinetuning} and \ref{tab:counting_as_image_sentence_aligment_finetuned}, we report our main results without and with additional finetuning on the counting training data, respectively.

\paragraph{No Finetuning}
As Table  \ref{tab:counting_as_image_sentence_aligment_nofinetuning} shows, accuracy for both ViLBERT and LXMERT is below or close to the random baseline, improving slightly on the baseline on pairwise accuracy.
We note that ViLBERT identifies correct image-sentence pairs relatively well ($73$--$79$ $p_c$), while failing on foils  ($24$--$27$ $p_f$).
This trend is also visible in ViLBERT 12-in-1; however all scores tend to improve when compared to ViLBERT (especially precision on foiled examples).
Roughly, we can rank models according to their performance from worse to best: ViLBERT, LXMERT, ViLBERT 12-in-1.
ViLBERT 12-in-1 is the only model that performs considerably above chance level according to standard accuracy when applied without direct finetuning (declarative format, standard and hard splits).
From these initial results, it seems that whereas ViLBERT 12-in-1 is able to identify correct image-sentence pairs well (i.e., up to $\sim 80$ $p_c$), its most important gains come from improved precision on foiled examples (up to $58$ $p_f$).

Overall, results when performing pairwise scoring ($acc_r$)
agree with the general trends in per-example results.
We can clearly observe that ViLBERT performs close to chance, LXMERT is somewhat better ($57$--$73$), and ViLBERT 12-in-1 performs best ($70$--$77$).
All models perform better when evaluated using pairwise accuracy compared to per-example accuracy. For example, ViLBERT 12-in-1 performs below chance level in per-example metrics ($43$ $acc$) but has good pairwise accuracy ($77$ $acc_r$). Thus, $acc_r$ is a less strict metric than standard accuracy $acc$.

\paragraph{With Finetuning}
In Table~\ref{tab:counting_as_image_sentence_aligment_finetuned} we note that when models are directly finetuned on counting training data, results tend to improve (except for the \textit{interpolated} data split, which we discuss separately further below).
ViLBERT improves overall accuracy ($acc$) on the standard split considerably to about $71$--$74$, but it still struggles on the hard and interpolated splits.
Results on the hard split are not good and have very high variance, which could be due to the model overfitting on the small amount of counting training data.
When finetuning ViLBERT 12-in-1 further on counting data, results also improve compared to the `no finetuning` setting.
As expected, ViLBERT 12-in-1 clearly outperforms ViLBERT on both standard and hard splits according to all metrics evaluated: on per-example metrics ($acc$, $p_c$, $p_f$) and also according to a pairwise ranking comparison ($acc_r$).

Pairwise results exhibit more consistent differences between splits, i.e., interpolated $\leq$ hard $\leq$ standard.
Both ViLBERT and ViLBERT 12-in-1 yield satisfying results of $76$--$83$ $acc_r$ in the standard and $56$--$77$ in the hard split.

\paragraph{Interpolated}
Finally, in our \textit{interpolated} split, we train on examples where correct answers are even numbers and test on examples where correct answers are odd numbers.
By doing that, we gain a glimpse into whether models are really learning to count, in which case interpolating even/odd numbers should be a relatively simple task.
We first note that models fail badly when finetuned and evaluated on interpolated data, achieving per-example accuracies between $48$--$61$ while still failing almost completely at identifying correct matches, as illustrated by precision $p_c$ close to zero.
Failure in the interpolated split is more clearly seen by inspecting pairwise accuracies, which are in the range $11$--$18$ and well below the random baseline of $50$.
Although ViLBERT 12-in-1 achieves reasonable results on the standard and hard splits, it still fails completely on the interpolated split.
This is in stark contrast to recent findings with  text-only pretrained language models, which have a good grasp of numeracy and perform well when interpolating quantities \cite{wallace-etal-2019-nlp}.

\subsubsection{Counting as Masked Language Modelling}\label{sec:MLM}

In this experiment, we set the image-sentence alignment head aside and employ the MLM capacity of LXMERT to further test its pretrained visual-linguistic representations on counting.

\begin{figure}
    \centering
    \includegraphics[width=.5\textwidth]{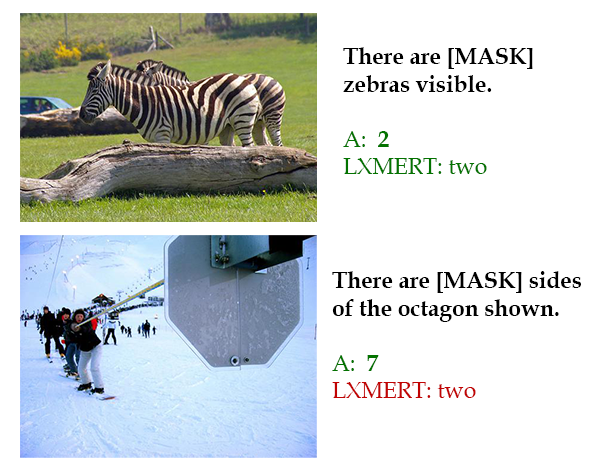}
    \caption{Two examples of applying masked language modelling on the counting dataset with LXMERT.}
    \label{fig:MLM_example}
\end{figure}

We mask the numeral in the declarative statements of our counting dataset and use LXMERT to predict the \texttt{[MASK]} token (see Figure \ref{fig:MLM_example}).
The model assigns probabilities to all words in its English vocabulary, comprising more than 30k words.
We remove vocabulary items that are neither numerals nor denote numerical quantities,\footnote{We include the indefinite article `a' and the negation `no' in our definition of a vocabulary item that denotes a numerical quantity, since they are interpretable as indicating `one' and `zero' respectively.} sort the remaining items in order of descending probability, and obtain a list of all numerical quantities LXMERT predicts for the masked token, ordered by likelihood.
In that list, we count the rank of the correct numeral in any formulation (\emph{e.g.} ``1'', ``one'', ``a'') and compute Recall@$k$ and mean rank (MR). We report the results in Table \ref{tab:mlm_LXMERT}, where we also show aggregate results per answer.

\begin{table}[t!]
    \small
    \centering
    \begin{tabular}{lrrrrrr}
        \bf Num. & \multicolumn{2}{c}{\bf Recall@1} & \multicolumn{2}{c}{\bf Recall@2} & \multicolumn{2}{c}{\bf MR}  \\
        & std. & hard & std. & hard & std. & hard \\
        \toprule
        \bf overall & 55.0 & 31.5 & 71.1 & 45.0 & 6.6 & 13.5 \\
        \midrule
        zero & 59.8 & 63.7 & 65.2 & 69.1 & 2.89 & 2.5 \\
        one & 86.2 & 86.6 & 89.7 & 89.0 & 1.6 & 1.7 \\
        two & 81.9 & 80.1 & 92.4 & 90.9 & 1.3 & 1.3 \\
        three & 15.6 & 12.9 & 85.2 & 87.1 & 2.0 & 2.0 \\
        four & 6.5 & 6.2 & 20.9 & 19.9 & 3.0 & 3.0 \\
        five & 0.4 & 0.0 & 2.5 & 2.9 & 5.1 & 5.1 \\
        six & 0.5 & 0.0 & 0.5 & 0.0 & 6.7 & 6.7 \\
        7 & 0.0 & 0.0 & 0.0 & 0.0 & 11.0 & 10.7 \\
        8 & 0.0 & 0.0 & 0.0 & 0.0 & 9.7 & 9.3 \\
        9 & 0.0 & 0.0 & 0.0 & 0.0 & 12.3 & 12.3 \\
        10 & 0.0 & 0.0 & 0.0 & 0.0 & 17.6 & 18.2 \\
        11 & 0.0 & 0.0 & 0.0 & 0.0 & 20.4 & 20.4 \\
        12 & 0.0 & 0.0 & 0.0 & 0.0 & 15.1 & 15.2 \\
        13 & 0.0 & 0.0 & 0.0 & 0.0 & 25.4 & 25.5 \\
        ... & ... & ... & ... & ... & ... & ... \\
        \bottomrule
    \end{tabular}
    \caption{Masked language modelling with LXMERT. 
    We report \textbf{Recall@$k$} and the mean rank (\textbf{MR}) of the predicted numeral on our counting dataset's test split.}
    \label{tab:mlm_LXMERT}
\end{table}

The results of the \emph{overall} Recall@$k$ and MR show clear differences between the standard and hard splits, whereas Recall@$k$ and MR \emph{per numeral} exhibit very consistent results for the same splits.
Recall that in the hard split numerals are more evenly distributed, whereas in the standard split the frequencies of different answers follow a more Zipfian distribution (Figure \ref{fig:counting_data_statistics}). 
This shows that the model has a strong preference for the numerals ``2'', followed by ``1'' and ``0'', suggesting that performance is largely determined by the statistical bias in the training data, rather than the specifics of the visual input in relation to the text.

Bias can also be observed when the MLM predictions are wrong.
While Table \ref{tab:mlm_LXMERT} reports metrics depicting how often the model is \emph{correct}, we further analyse the cases where model predictions are \emph{wrong}.
Among image-sentence pairs in the hard split where the model prediction is wrong, the model predicts: ``two'' 51\% of the time, ``no'' 12\% of the time (which we count as ``0''), followed by ``three'' (12\%), ``four'' (5\%) and ``five'' (1\%).

Good performance on low quantities reflects their frequent occurrence in V\&L datasets \citep{goyal2017making}. The very poor performance on under-represented quantities suggests both a lack of generalisation of the V\&L model, as well as limitations arising from models' Faster R-CNN \citep{ren2015faster} visual backbone.

\begin{table}[t!]
    \small
    \centering
    \resizebox{\linewidth}{!}{
    \begin{tabular}{lrr | lrr}
        \bf Answer & \multicolumn{2}{c|}{\bf VQA accuracy} & \bf Answer & \multicolumn{2}{c}{\bf VQA accuracy}\\
        & std. & hard && std. & hard \\
        \toprule
        \bf overall & $\bm{53.9}$ & $\bm{41.8}$ & seven & 4.0 & 3.8 \\
        zero & 94.4 & 93.7 & eight & 18.0 & 14.5 \\
        one & 75.0 & 69.7 & nine & 3.6 & 2.5 \\
        two & 62.0 & 62.4 & ten & 12.0 & 11.6 \\
        three & 32.8 & 31.4 & eleven & 10.0 & 8.7 \\
        four & 25.0 & 21.3 & twelve & 28.6 & 32.3\\
        five & 15.6 & 17.1 & thirteen & 12.5 & 7.1 \\
        six  & 17.4 & 19.4 &  ... & ... & ... \\
        \bottomrule
    \end{tabular}
    }
    \caption{Overall (in bold) and per-answer accuracy of LXMERT further fine-tuned on the VQA task \citep{antol2015vqa} on the standard and hard counting splits.}
    \label{tab:VQA_LXMERT}
\end{table}

\subsubsection{Counting as VQA}\label{sec:LXMERT_VQA}
Finally, we frame the counting probe in its standard setting as a VQA problem, without foiling.
We use the publicly available LXMERT model further finetuned on the VQA v2.0 dataset \citep{goyal2017making}.
The test setting is the same as for the original VQA task: the model receives questions (``How many ...?'') from our counting dataset as input and has to predict the most likely answer from a list of 3,129 possible answers.
All answers in our dataset are contained in the model's answer list.
We report detailed results in Table \ref{tab:VQA_LXMERT}.

The model achieves an overall 53.9 accuracy on the standard split and 41.8 on the hard split. The detailed accuracies \emph{per numeral} show the same trend as Table \ref{tab:mlm_LXMERT}.
Differences in performance between the standard (Zipfian) and the hard (more balanced) splits are predominantly due to the different proportion of quantities in the splits.
Once again, this reveals a lack of generalisation coupled with a surplus of bias exploitation potential: the model relies on highly frequent quantities like ``one'' or ``two'' as a ``safe bet'' when predicting under uncertainty.

A notable difference between using LXMERT's MLM head without direct finetuning (as in Table \ref{tab:mlm_LXMERT}), and using LXMERT's VQA head further finetuned on VQA v2.0 (Table \ref{tab:VQA_LXMERT}) is seen for the numeral ``zero'': the model's capacity to predict ``zero'' is enhanced by the finetuning process, to the detriment of the frequent quantities ``one'' and ``two''. Fine-tuning also seems to improve prediction for numerals ``four'' to ``six'', and also for ``12''.\footnote{``12'' or ``a dozen'' is a frequent answer in VQA v2.0.}
Counting further than that is a challenge for LXMERT.



\section{Related Work}\label{sec:related}
Originally proposed for text-only models \cite{devlin2019bert,wang2019_superglue,lewis-etal-2020-bart}, the \textit{pretrain-and-finetune} paradigm has become the \textit{de facto} standard for vision and language tasks.
The core idea is that pretraining on large and diverse datasets should lead to robust multimodal representations, so that models can be easily finetuned for different tasks.

\paragraph{Pretrained Vision \& Language Models}
Based on the \textit{pretrain-and-finetune} paradigm, many pretrained V\&L models have recently been proposed which combine images and text using BERT-like architectures. They include ViLBERT \cite{lu2019vilbert,lu2020vilbert12in1}, LXMERT \cite{tan-bansal-2019-lxmert}, VisualBERT \cite{li2019visualbert}, UNITER \cite{chen2020uniter}, Unicoder-VL \cite{Li-etal-2020unicodervl}, VL-BERT \cite{Su2020VL-BERT}, among others.
They can be classified into \textit{single-} or \textit{dual-stream} architectures:
single-stream models concatenate words to object bounding box features and encode this sequence using a single transformer stack;
dual-stream models have separate transformer stacks for visual and textual inputs, with layers to fuse these into multimodal features (i.e., co-attention layers).
ViLBERT, ViLBERT 12-in-1, and LXMERT, i.e. the models we use with in this work, are all dual-stream.

\citet{bugliarello2020multimodal} find that single- and dual-stream models perform comparably under similar conditions.
\citet{ilharco2020probing} show that contextual text-only language models such as BERT encode visual representations reasonably well, though they fall short of human performance.
In the context of the VALUE benchmark, \citet{cao2020behind} report results on multiple V\&L tasks, some of which we also corroborate, notably, the dominance of textual features compared to visual features in the model's predictions.

\paragraph{Vision \& Language Models for Counting}
Counting is known to be hard for V\&L models, and has been studied extensively
\cite{segui2015learning,Chattopadhyay-etal-2017,trott2018interpretable,zhang2018learning,Acharya_Kafle_Kanan_2019TallyQA}. 

\citet{Chattopadhyay-etal-2017} investigate strategies based on object detection, regression, subitising and averaging over the results returned by a model ensemble.
\citet{trott2018interpretable} create the \textit{HowMany-QA} counting dataset.
Their Interpretable RL Counter (IRLC) model solves counting by iteratively including objects in a pool, whose size is then reported.
\citet{Acharya_Kafle_Kanan_2019TallyQA} propose \textit{TallyQA}, a large counting dataset which includes both simple questions (e.g. {\em How many giraffes?}) and harder cases involving additional properties (e.g. {\em How many giraffes \underline{are sitting down?}}).
Finally, \citet{zhang2018learning} argue that attention bottlenecks compromise counting capabilities \cite[see][Section 3]{zhang2018learning}, showing that an alternative architecture which includes a branch specifically designed to overcome the bottleneck for counting leads to considerable improvements.


\paragraph{Counting and the attention bottleneck}
The `attention bottleneck' noted by \citet{zhang2018learning} and further discussed by \citet{Acharya_Kafle_Kanan_2019TallyQA} generally afflicts architectures where the image pipeline has the general form ``image $\rightarrow$ CNN $\rightarrow$  convolutional feature maps $\rightarrow$  attention bottleneck $\rightarrow$  prediction''. The `bottleneck' is created by the attention mechanism between input and prediction layer. For details and examples, see
Appendix \ref{sec:appendix:attention_bottleneck}.

This issue does not apply to the pretrained V\&L models reviewed above, or the models we experiment with.
ViLBERT, ViLBERT 12-in-1, and LXMERT have two multi-layer transformer stacks to encode image and text, respectively. 
None of these models have an attention bottleneck; rather, the outputs of modality-specific encoders are integrated via multiple co-attention layers.
When finetuning the model on a target task, a prediction head is commonly trained from scratch and uses the output of the last co-attention layer as input.

\section{Conclusions and Future Work}\label{sec:conclusion}

We probed three pretrained V\&L models on image-sentence alignment and counting: two tasks that require joint understanding of image, text and their correspondence.
Our results show image-text alignment capabilities which range from good (for ViLBERT and\mbox{ViLBERT 12-in-1}) to satisfactory (for LXMERT).
Our results highlight that LXMERT (and to a lesser extent, ViLBERT 12-in-1) may be suffering from catastrophic forgetting. As for counting, we observe sub-optimal performance in all models investigated, even after finetuning on counting data. In these models, there is limited evidence of grounding of symbols in visual data after pretraining; all models exploit biases in the data and seem to lack the capability to individuate entities in the visual input, a prerequisite for counting.
Our results raise concerns about heavyweight V\&L models, whose main selling point is their ability to solve complex tasks.
Our findings suggest that understanding their capabilities requires more targeted investigations on specific phenomena. In line with this reasoning, our ongoing work is aiming towards a benchmark that will address several linguistic phenomena in addition to counting. We hope such a benchmark will serve the community to probe the grounding capabilities of vision and language models on a broad range of linguistic phenomena.

More generally, we encourage researchers i) to report the performance on pretraining tasks,
ii) to work towards effective pretraining, and iii) to test for catastrophic forgetting during finetuning. 
The high computational and environmental cost of current pretraining practices may outweigh the benefits of reusing such models, 
leaving the prospect of lightweight and green AI as a distant goal.

\section*{Acknowledgments}
IC has received funding from the European Union’s Horizon 2020 research and innovation programme under the Marie Sk\l{}odowska-Curie grant agreement No 838188. 
AG is supported by the European Union's Horizon 2020 research and innovation Programme under the Marie Sk\l{}odowska-Curie grant agreement No 860621. 
This collaboration was facilitated by the Multi3Generation COST Action CA18231.

\bibliography{anthology,eacl2021}
\bibliographystyle{acl_natbib}

\clearpage
\appendix
\section{Appendix}\label{sec:appendix}

\subsection{Training and evaluation setup}\label{sec:appendix:training_setup}
In our experiments with ViLBERT and ViLBERT 12-in-1, we use the \url{https://github.com/facebookresearch/vilbert-multi-task} codebase.
In our ``zero-shot'' experiments, we use ViLBERT pretrained on image-sentence ranking\footnote{\url{https://dl.fbaipublicfiles.com/vilbert-multi-task/pretrained_model.bin}} and ViLBERT 12-in-1 pretrained on twelve tasks.\footnote{\url{https://dl.fbaipublicfiles.com/vilbert-multi-task/multi_task_model.bin}}
In our experiments with LXMERT, we use the \url{https://github.com/huggingface/transformers} codebase.
For evaluating LXMERT on image-sentence alignment or counting as MLM, we use the publicly available pretrained model.\footnote{\texttt{LxmertForPreTraining.from\_pretrained} and model name ''unc-nlp/lxmert-base-uncased''.}
When evaluating LXMERT on counting as VQA, we use the publicly available model additionally fine-tuned on the VQA 2.0 dataset \cite{goyal2017making}.\footnote{\texttt{LxmertForQuestionAnswering.from\_pretrained} and model name ''unc-nlp/lxmert-vqa-uncased''.}

When finetuning ViLBERT and \mbox{ViLBERT 12-in-1} on our counting dataset, we train models on the training split and evaluate on the concatenation of the validation and test splits (see Table \ref{tab:counting_data_statistics} for details on the splits).
We train all models for 20 epochs and evaluate always at the end of each epoch, therefore 20 times. For each model, we report the best scores obtained across all 20 evaluations.
ViLBERT and ViLBERT 12-in-1 are finetuned on our counting data following the standard finetuning procedure of ViLBERT 12-in-1: AdamW optimiser \citep{loshchilov2018decoupled} with a learning rate 4e-5 and a linear warm-up scheduler, batch size 16, and a maximum of 100 detected objects per image, a text backbone \textit{bert-base-uncased} and configuration file \textit{bert\_base\_6layer\_6conect.json}.
We finetune models using the binary cross-entropy loss where the task is to decide if an image-sentence pair is correct or a foil, and each instance consists of a question (or statement, see Section \ref{sec:appendix:counting_template} below) about the number of objects in the image and an answer (that might be correct or foiled).

\subsection{Question-to-statement template}\label{sec:appendix:counting_template}

We create a few simple templates to convert \texttt{<}question\texttt{,} answer\texttt{>} pairs into a declarative statement.
We denote the answer as \texttt{A}, and by definition it is always a number.
Other capitalised letters (e.g., \texttt{B}, \texttt{C}, etc.) denote entire sets of words that are either copied over to the declarative sentence or removed according to the template.
If a set of words is optional in the template, it is enclosed in brackets, e.g. \texttt{[D]}.
A template is selected if there is substring match between the \textit{template's key} and the question.
We denote negation by $\sim$.
We process templates in order so that if a template matches, it ``consumes'' the QA pair and produces a declarative sentence.
If no template matches, the QA pair is ignored and not added to our counting dataset.

\paragraph{``are there''} \texttt{How many B are there [C]?  $\rightarrow$ There are A B [C].} E.g.: ``How many black cats are there in the picture?'' $\rightarrow$ ``There are \texttt{A} black cats in the picture.''

\paragraph{``can you see''} \texttt{How many B can you see [C]?  $\rightarrow$ You see A B [C].} E.g.: ``How many elephants can you see?'' $\rightarrow$ ``You see \texttt{A} elephants.''

\paragraph{``do you see''} \texttt{How many B do you see [C]?  $\rightarrow$ There are A B [C].} E.g.: ``How many people do you see by the tree?'' $\rightarrow$ ``There are \texttt{A} people by the tree.''

\paragraph{``are''} \texttt{How many B are C?  $\rightarrow$ There are A B C.} E.g.: ``How many glasses are on the table?'' $\rightarrow$ ``There are \texttt{A} glasses on the table.''

\paragraph{``can''} \texttt{How many B can C?  $\rightarrow$ A B can C.} E.g.: ``How many surcoats can be found in the storage?'' $\rightarrow$ ``\texttt{A} surcoats can be found in the storage.''

\paragraph{``do'' and ``have''} \texttt{How many B do C have [D]?  $\rightarrow$ C have A B [D].} E.g.: ``How many headphones do the people have?'' $\rightarrow$ ``The people have \texttt{A} headphones.''

\paragraph{``does'' and ``have''} \texttt{How many B does C have D?  $\rightarrow$ C has A B C.} E.g.: ``How many holes does he have in his pants?'' $\rightarrow$ ``He has \texttt{A} holes in his pants.''

\paragraph{``have''} \texttt{How many B have C?  $\rightarrow$ A B have C.} E.g.: ``How many bottles have blue caps?'' $\rightarrow$ ``\texttt{A} bottles have blue caps.''

\paragraph{$\sim$ ``is'' and $\sim$ ``will'' and $\sim$ ``does'' and $\sim$ ``has''} \texttt{How many B?  $\rightarrow$ There are A B.} E.g.: ``How many cars in the picture?'' $\rightarrow$ ``There are \texttt{A} cars in the picture.''

\subsubsection{Plurals}
Finally, after applying the above mentioned templates we check if the original answer to the question is the number 1.
When that is the case, we convert all sentences starting with ``There are'' by \texttt{There are B.} $\rightarrow$ \texttt{There is B.}
We also transform the following words: ``people'' $\rightarrow$ ``person'', ``men'' $\rightarrow$ ``man'', ``women'' $\rightarrow$ ``woman'', and also remove the final ``s'' of words up to the fourth word in the declarative sentence (all words but ``has'').

\subsection{Counting Data}\label{sec:appendix:counting_data}
\begin{table}
    \small
    \centering
    \begin{tabular}{llrrr}
        \bf Split & & \bf \#Train & \bf \#Valid & \bf \#Test \\
        \toprule
        \multirow{3}{*}{Standard} & \bf Correct & $6,001$ & $2,439$ & $3,622$ \\
        & \bf Foiled & $17,896$ & $7,283$ & $10,800$ \\
        \cmidrule{2-5}
        & \bf Total & $23,897$ & $9,722$ & $14,422$ \\
        \midrule
        \multirow{3}{*}{Hard} & \bf Correct & $1,545$ & $1,130$ & $1,352$ \\
        & \bf Foiled & $4,610$ & $3,378$ & $4,040$ \\
        \cmidrule{2-5}
        & \bf Total & $6,239$ & $4,508$ & $5,392$ \\
        \midrule
        \multirow{3}{*}{Interpolated} & \bf Correct & $3,303$ & $1,331$ & $2,013$ \\
        & \bf Foiled & $9,840$ & $3,969$ & $5,998$ \\
        \cmidrule{2-5}
        & \bf Total & $13,143$ & $5,300$ & $8,011$ \\
        \bottomrule
    \end{tabular}
    \caption{Counting data statistics.}
    \label{tab:counting_data_statistics}
\end{table}

In Table \ref{tab:counting_data_statistics} we show the statistics in our counting datasets. 

We note that: the \textit{hard} split has considerably fewer examples than the other two splits, due to the capping at $k=200$ examples per answer type; furthermore, the \textit{interpolated} split also 
has fewer examples than the \textit{standard} split because we discard all examples with odd answers from its training set and all examples with even answers from its validation and test sets.

\begin{table}[t!]
    \small
    \centering
    \begin{tabular}{l rrrrrr}
        \bf Numeral & \multicolumn{6}{c}{\bf Percentage (\%)} \\
        & \multicolumn{2}{c}{\bf Train} & \multicolumn{2}{c}{\bf Valid} & \multicolumn{2}{c}{\bf Test} \\
        & std. & hard & std. & hard & std. & hard \\
        \toprule
        zero & 9 & 11 & 7 & 15 & 7 & 12 \\
        one & 30 & 11 & 26 & 14 & 27 & 12 \\
        two & 25 & 9 & 32 & 15 & 19 & 13 \\
        three & 14 & 11 & 13 & 13 & 16 & 12 \\
        four & 8 & 9 & 9 & 15 & 7 & 12 \\
        five & 5 & 12 & 4 & 10 & 5 & 13 \\
        six & 3 & 11 & 3 & 6 & 3 & 9 \\
        7 & 1 & 5 & 1 & 2 & 1 & 4 \\
        8-10    & 3 & 11 & 2 & 5 & 3 & 8 \\
        10-20   & 2 & 7 & 2 & 4 & 2 & 5 \\
        21+     & 0 & 0 & 0 & 1 & 0 & 1 \\
        \bottomrule
    \end{tabular}

    \caption{Percentage of numerals in the counting data.}
    \label{tab:numeral_frequencies}
\end{table}

The \textit{hard} split is more balanced with regards to the number of classes, whereas quantities in the \emph{standard} split follow a more natural distribution, where numerals like ``one'', ``two'' or ``three'' are more common than large quantities or mentions of empty sets (see Figure~\ref{fig:counting_data_statistics}). This more skewed distribution is made even more evident in Table \ref{tab:numeral_frequencies}, which shows the percentage of occurrence of numerals in the standard split.
The less skewed distribution in the \emph{hard} split would be expected to be harder, since we artificially lower the relative frequency of frequent answers (compare the inner to the outer circles in Figure \ref{fig:counting_data_statistics}).


\subsection{Counting and the attention bottleneck}\label{sec:appendix:attention_bottleneck}
The attention bottleneck takes place when there is an image encoder model and there is a bottleneck between the model input and the layer that makes the predictions of interest.
This situation can be exemplified where the image pipeline has the general form ``image $\rightarrow$ CNN $\rightarrow$  convolutional feature maps $\rightarrow$  attention bottleneck $\rightarrow$  prediction''.
We now use an idealised example meant to illustrate the attention bottleneck issue, similar to the one used in \citet{zhang2018learning}.
The goal is 
to clarify when the issue should arise and in what conditions.

Imagine there is a \textit{cat prediction} model, and we present it with an image with a single cat.
After a number of CNN layers, the model computes a convolutional feature map $\bm{c}_\text{i,j}$.
In the attention bottleneck, a perfectly trained model will assign probability close to 1 to the ``cat'' feature vector, e.g., say feature map $\bm{c}_\text{4,7}$, and $0$ elsewhere, and the attention output will roughly be $1 \cdot \bm{c}_\text{4,7} + 0 \cdot \sum_{i \neq 4, j \neq 7}c_{i,j}$.

We can now think of an idealised scenario where we create an identical copy of the cat image and paste it side-by-side with the original image (so that there are two cats), or we can think of another image which depicts two identically looking cats. 
When encoding any such image, each ``cat'' feature vector should get $\sim 0.5$ probability in the attention layer, again assuming an idealised and perfectly trained model.
By design of the attention mechanism, the attention output will consist of the two sets of ``cat'' features multiplied by $\sim 0.5$ each and summed together.
Therefore the attention output for the image with two cats would virtually be indistinguishable from the output for the image with a single cat.
If the model only has access to these features, i.e., the attention mechanism is a bottleneck, it becomes very hard for the model to count, which by definition would require being able to differentiate 
the number of cats in the input image.

\end{document}